\documentclass[preprint,12pt]{elsarticle}

\usepackage{caption}
\usepackage{times,amsmath,epsfig}
\usepackage{epstopdf}
\usepackage{color}
\usepackage{graphicx}  
\usepackage{cases}
\usepackage{algorithm}
\usepackage{algorithmic}  
\usepackage{textcomp} 
\usepackage{tikz}
\usepackage{amsthm}
\usepackage{arydshln}
\usepackage[hidelinks]{hyperref}
\usepackage{flushend}

\usepackage[utf8]{inputenc}
\usepackage[english]{babel}

\theoremstyle{plain}

\makeatletter
  \newcommand*{\rom}[1]{\expandafter\@slowromancap\romannumeral #1@}
\makeatother

\usepackage{enumerate}
\usepackage{amsmath,amssymb,amsfonts}
\usepackage{algorithmic}
\usepackage{graphicx}
\usepackage{textcomp}
\usepackage{color}
\usepackage[normalem]{ulem}
\useunder{\uline}{\ul}{}
\def\BibTeX{{\rm B\kern-.05em{\sc i\kern-.025em b}\kern-.08em
    T\kern-.1667em\lower.7ex\hbox{E}\kern-.125emX}}

\def\0{{\mathbf 0}}
\def\1{{\mathbf 1}}

\def\h{{\mathbf h}}

\def\l{{\mathbf l}}

\def\p{{\mathbf p}}
\def\q{{\mathbf q}}

\def\v{{\mathbf v}}

\def\x{{\mathbf x}}

\def\A{{\mathbf A}}

\def\D{{\mathbf D}}

\def\I{{\mathbf I}}

\def\L{{\mathbf L}}

\def\P{{\mathbf P}}

\def\W{{\mathbf W}}

\def\ie{{\textit{i.e.}}}

\def\cE{{\mathcal E}}

\def\cG{{\mathcal G}}

\def\cN{{\mathcal N}}

\def\cV{{\mathcal V}}

\def\0{{\mathbf 0}}
\def\1{{\mathbf 1}}

\def\h{{\mathbf h}}

\def\l{{\mathbf l}}

\def\p{{\mathbf p}}
\def\q{{\mathbf q}}

\def\v{{\mathbf v}}

\def\x{{\mathbf x}}

\def\A{{\mathbf A}}

\def\D{{\mathbf D}}

\def\I{{\mathbf I}}

\def\L{{\mathbf L}}

\def\P{{\mathbf P}}

\def\W{{\mathbf W}}

\def\ie{{\textit{i.e.}}}

\def\cE{{\mathcal E}}

\def\cG{{\mathcal G}}

\def\cN{{\mathcal N}}

\def\cV{{\mathcal V}}

\def\cN{{\mathcal N}}

\def\etal{\textit{et al.}}

\newcommand\norm[1]{\left\lVert#1\right\rVert}

\usepackage{multirow}
\usepackage{algorithm}
\usepackage{algorithmic}

\usepackage{amssymb}

\journal{}

\begin{document}

\begin{frontmatter}



\title{Understanding Key Point Cloud Features for Development Three-dimensional Adversarial Attacks}





\author[inst1]{Hanieh Naderi}
\affiliation[inst1]{organization={College of Interdisciplinary Science and Technologies},
            addressline={University of Tehran}, 
            city={Tehran},
            country={Iran}}


\author[inst2,inst3]{Chinthaka Dinesh}
\affiliation[inst2]{organization={Northeastern University, Vancouver},
            city={BC},
            country={Canada}}

\author[inst3]{Ivan V. Baji\'{c}}
\affiliation[inst3]{organization={School of Engineering Science},
            addressline={Simon Fraser University}, 
            city={BC},
            country={Canada}}            

\author[inst4]{Shohreh Kasaei}
\affiliation[inst4]{organization={Department of Computer Engineering},
            addressline={Sharif University of Technology}, 
            city={Tehran},
            country={Iran}}



\begin{abstract}
Adversarial attacks pose serious challenges for deep neural network (DNN)-based analysis of various input signals. In the case of three-dimensional point clouds, methods have been developed to identify points that play a key role in network decision, and these become crucial in generating existing adversarial attacks. For example, a saliency map approach is a popular method for identifying adversarial drop points, whose removal would significantly impact the network decision. This paper seeks to enhance the understanding of three-dimensional adversarial attacks by exploring which point cloud features are most important for predicting adversarial points. 
Specifically, Fourteen key point cloud features such as edge intensity and distance from the centroid are defined, and multiple linear regression is employed to assess their predictive power for adversarial points. Based on critical feature selection insights, a new attack method has been developed to evaluate whether the selected features can generate an attack successfully.  Unlike traditional attack methods that rely on model-specific vulnerabilities, this approach focuses on the intrinsic characteristics of the point clouds themselves. It is demonstrated that these features can predict adversarial points across four different DNN architectures— Point Network (PointNet), PointNet++, Dynamic Graph Convolutional Neural Networks (DGCNN), and Point Convolutional Network (PointConv)—outperforming random guessing and achieving results comparable to saliency map-based attacks. This study has important engineering applications, such as enhancing the security and robustness of three-dimensional point cloud-based systems in fields like robotics and autonomous driving.
\end{abstract}


\end{frontmatter}


\section{Introduction}
\label{sec:intro}
DNNs have become a go-to approach for many problems in image processing and computer vision~\cite{denoising2017tip,dehazing2023tip,inverse2017tip,hyperspectral2022tip,naderi2020scale} due to their ability to model complex input-output relationships from a relatively limited set of data. However, studies have also shown that DNNs are vulnerable to adversarial attacks~\cite{goodfellow2014explaining,szegedy2013intriguing}. An adversarial attack involves constructing an input to the model (adversarial example) whose purpose is to cause the model to make a wrong decision. 
Much literature has been devoted to the construction of Two-Dimensional (2D) adversarial examples for image analysis models and the exploration of related defenses~\cite{moosavi2017universal,carlini2017towards,goodfellow2014explaining,szegedy2013intriguing,naderi2022generating}. Research on adversarial attacks and defenses has gradually expanded to Three-Dimensional (3D) point cloud models as well, especially point cloud classification~\cite{liu2019extending,xiang2019generating,huang2022shape,wen20233d,naderi2023survey,liu2023point,huang2024pointcat,zhang2023ada3diff,hamdi2020advpc}. 

Point clouds themselves have become an increasingly important research topic \cite{pra-net2021tip,app-net2023tip,raht2016tip,pcdenoising2020tip}. Given a deep model for point cloud classification, a number of methods have been proposed to determine critical points that could be used in an adversarial attack~\cite{zheng2019pointcloud,zhang2019explaining,fan2022salient}. For example, Zheng \etal~\cite{zheng2019pointcloud} proposed a differentiable method of shifting points to the center of the cloud, known as a \textit{saliency map technique}, which approximates point dropping and assigns contribution scores to input points based on the resulting loss value. Other methods to determine critical points similarly try to estimate the effect of point disturbance on the output.  

Once the critical points have been determined, they can be used to create adversarial examples. 
Several recent studies~\cite{liu2019extending,yang2021adversarial,xiang2019generating} use critical points as initial positions and then introduce perturbations to create attacks. Usually, some distance-related criteria, such as Hausdorff~\cite{xiang2019generating} or the Chamfer distance~\cite{xiang2019generating}, are used to constrain perturbations around critical positions. 
Instead of perturbation, another kind of attack drops the critical points; for example, the well-known Drop100 and Drop200 attacks~\cite{zheng2019pointcloud} drop, respectively, 100 and 200 points from the point cloud in order to force the model to make a wrong decision. These are considered to be among the most challenging attacks to defend against~\cite{wu2020if,naderi2023lpf,liu2019extending,zhou2019dup}.

The methods mentioned above, and others in the literature, require access to the DNN model in order to determine critical points. For example, ``white-box'' attacks have access to the model's internal architecture and parameters, while ``black-box'' attacks are able to query the model and obtain its output, but without the knowledge of internal details~\cite{naderi2023survey}. These approaches are in line with the popular view in the literature~\cite{szegedy2013intriguing}: that the existence of adversarial examples is a \textit{flaw of the model}, that they exist because the model is overly parametrized, nonlinear, etc. According to this reasoning, each model has its own flaws, i.e., its own critical points. Another view is that adversarial examples are consequences of the data distribution on which the model is trained~\cite{ilyas2019adversarial_not_bugs}. This would suggest that different models trained on the same data may share some adversarial examples, but they have to be determined in the context of the data distribution.

In this paper, a different point of view is presented, demonstrating that critical points in point clouds can be determined from the features of the point cloud itself. To our knowledge, this is the first work in the point cloud literature to explore which features play a key role in adversarial attacks. In a broader context, we suggest that critical points in point clouds are inherent characteristics of the point clouds themselves, which give them their crucial properties that are important in their analysis, i.e., what makes an airplane - an airplane, or a chair - a chair. 

In addition to the novel point of view, a new attack has been created based on the methodology, leading to a less computationally expensive way of generating adversarial attacks compared to other methods.

Furthermore, the attack has the potential to generalize better and be more transferrable to different models.

The rest of the paper is organized as follows. The related work is discussed in Section~\ref{sec:related}, together with a more detailed explanation of our contribution. Our proposed methodology is presented in Sections~\ref{sec:features} and~\ref{sec:Proposed_Method}.  
The experimental results are reported in Section~\ref{sec:results}, followed by conclusions in Section~\ref{sec:conclude}.

\section{Related Work}
\label{sec:related}
\subsection{Deep models for point cloud analysis}
PointNet~\cite{qi2017pointnet} was a pioneering approach for DNN-based point cloud analysis. Learnt features are extracted from individual points in the input point cloud and then aggregated to global features via max-pooling. 
As a result of these global features, a shape can be summarized by a sparse set of key points, also called the \emph{critical point set}. The authors of PointNet showed that any set of points between the critical point set and another set called the \emph{upper bound shape} will give the same set of global features, and thus lead to the same network decision. While this proved a certain level of robustness of PointNet to input perturbations, it also pointed to strong reliance on the critical point set, 
which was subsequently used to design various adversarial attacks. 

PointNet has inspired much subsequent work on DNN-based point cloud analysis, of which we review only three approaches subsequently used in our experiments.  
One of these is PointNet++~\cite{qi2017pointnet1}, a hierarchical network designed to capture fine geometric structures in a point cloud. Three layers make up PointNet++: the sampling layer, the grouping layer, and the PointNet-based learning layer. These three layers are repeated in PointNet++ to learn local geometric structures.
Another representative work is Dynamic Graph Convolutional Neural Network (DGCNN)~\cite{wang2019dynamic}. It exploits local geometric structures by creating a local neighborhood graph and using convolution-like operations on the edges connecting neighboring pairs of points. PointConv~\cite{wu2019pointconv} is another architecture that extends PointNet by incorporating convolutional layers that work on 3D point clouds. To better handle local features in point cloud data, a multi-layer perceptron (MLP) is used to approximate weight functions, and inverse density scale is used to re-weight these functions.

\subsection{Adversarial attacks on point clouds}
Point clouds are defined by the 3D coordinates of points making up the cloud. Thus, adversarial attacks can be performed by adding, dropping, or shifting points in the input cloud. An adversarial attack can be created by examining all points in the input cloud, or just critical points as potential targets.
Liu \etal~\cite{liu2019extending} were inspired by the success of gradient-guided attack methods, such as Fast Gradient Sign Method (FGSM)~\cite{goodfellow2014explaining} and Projected Gradient Descent (PGD)~\cite{madry2017towards}, on 2D images. They applied a similar methodology to develop adversarial attacks on 3D point clouds. 
Similarly, the Carlini and Wagner (C\&W) \cite{carlini2017towards} optimization for finding adversarial examples has also been transplanted to 3D data. For example, Tsai \etal~\cite{tsai2020robust} use the C\&W optimization formulation with an additional perturbation-bound regularization to construct adversarial attacks. 
To generate an attack with a minimum number of points, Kim \etal~\cite{kim2021minimal} extend the C\&W  formulation 
by adding a term 
to constrain the number of perturbed points. The adversarial points found in~\cite{kim2021minimal} 
were almost identical to the PointNet critical points.

Xiang \etal~\cite{xiang2019generating} demonstrated that PointNet can be fooled by shifting or adding synthetic points or adding clusters and objects to the point cloud. To find such adversarial examples, they applied the C\&W strategy to the critical points, rather than all points. Constraining the search space around critical points is sometimes necessary because an exhaustive search through an unconstrained 3D space is infeasible.
An attack method that uses the critical-point property for PointNet is proposed by Yang \etal~\cite{yang2021adversarial}. 
By recalculating the class-dependent importance for each remaining point, they iteratively remove the most crucial point 
for the true class. 
The authors noted that the critical points exist in different models and that a universal point-dropping method should be developed for all models.
Wicker \etal~\cite{wicker2019robustness} proposed randomly and iteratively determining the critical points and then generating adversarial examples by dropping these points. 

Arya \etal~\cite{arya2021adversarial} identify critical points by calculating the largest magnitudes of the loss gradient with respect to the points. After finding those points, the authors propose a minimal set of adversarial points among critical points and perturb them slightly to create 
adversarial examples. 
Zheng \etal~\cite{zheng2019pointcloud} developed a more flexible method that extends finding critical points to other deep models besides PointNet. They introduced a \emph{saliency score} defined as 
\begin{equation}
    s_i = -r_i^{1+\gamma} \frac{\partial \mathcal{L}}{\partial r_i},
    \label{eq:saliency_score}
\end{equation}
where $r_i$ is the distance of the $i$-th point to the cloud center, $\gamma$ is a hyperparameter, and $\frac{\partial  \mathcal{L}}{\partial r_i}$ is the gradient of the loss $\mathcal{L}$ with respect to the amount of shifting the point towards the center. Adversarial examples are created by shifting the points with high saliency scores towards the center, so that they will not affect the surfaces much.  
 
In addition to the methods for creating adversarial attacks on point clouds, a number of methods for defending against these attacks have been developed~\cite{naderi2023survey}. For 3D point cloud classification, adversarial training and point removal as a pre-processing step in training have been extensively studied~\cite{liu2020adversarial}. Some of the methods proposed for point removal to improve robustness against adversarial attacks include simple random sampling (SRS)~\cite{yang2021adversarial}, statistical outlier removal (SOR)~\cite{zhou2019dup}, Denoiser and UPsampler Network (DUP-Net)~\cite{zhou2019dup}, high-frequency removal~\cite{naderi2023lpf}, and salient point removal~\cite{liu2019extending}.

\subsection{Explainability Methods}

Explainability of 3D point cloud deep models is an important emerging area of research. Zhang~\etal~\cite{zheng2019pointcloud} introduced a class-attentive response map to visualize activated regions in PointNet, while later work~\cite{zhao2020evaluation} focused on interpreting 3D CNNs using statistical methods to evaluate convolution functions.

Nethod in~\cite{tayyub2022explaining} proposed iterative heatmaps to explain point cloud models, and Atkinson \etal~\cite{arnold2022improved} introduced a novel classification method that enhances explainability by integrating multiple layers of human-interpretable insights.
Other notable approaches include PointMask~\cite{taghanaki2020pointmask}, which used mutual information to mask points, and PointHop~\cite{zhang2020pointhop}, which applied local-to-global attributes for classification. These methods typically focus on extracting or visualization key point cloud features specific to a particular model, such as PointNet, to improve understanding and explainability. On the other hand, our approach seeks to identify key point cloud features derived from the data's intrinsic characteristics, making them generalizable across different models.

\subsection{Our contribution}
This work focuses on identifying key point cloud features, derived from intrinsic characteristics, that are crucial to a model’s decision-making process. 
To achieve this, We start by defining a set of fourteen point cloud features (Section~\ref{sec:features}) derived using concepts from graph signal processing, which represent various characteristics of a point cloud. 
To identify the critical point cloud features that influence a model's decision, focus is placed on adversarial drop attacks, with the specific goal of identifying \textit{adversarial drop points} whose removal is likely to change the target model's decision.

By applying multiple linear regression analysis, we examine which of those features are indicative of adversarial drop points identified using conventional methods (Section~\ref{sec:Proposed_Method}). Specifically, we examine three DNN models -- PointNet~\cite{qi2017pointnet}, PointNet++~\cite{qi2017pointnet1}, and DGCNN~\cite{phan2018dgcnn} -- and identify a set of features that have statistically significant influence in predicting adversarial drop points for these models. It is then demonstrated that a new attack can be created using a combination of these features derived directly from the point cloud. While our attack cannot be more successful than a black-box or white-box attack tailored to a given model, it is shown to be significantly more effective than a random attack. Additionally, the transferability of the developed attack to an unseen model, PointConv~\cite{wu2019pointconv}, is studied, and its effectiveness against different defenses is examined.

\section{Point Cloud Features}
\label{sec:features}
In this section, leveraging recent advances in \textit{graph signal processing} (GSP) \cite{shuman2013,liu2017}, a set of fourteen features is developed to represent various characteristics of a point cloud. These features will later be analyzed (Section~\ref{sec:Proposed_Method}) in terms of their ability to predict adversarial drop points. First, the basic concepts in GSP and the graph construction for a given 3D point cloud are reviewed, leading to the computation of graph-based features.

\subsection{Preliminaries}
\subsubsection{Graph definitions}
A undirected weighted graph $\cG~=~(\cV,\cE,\W)$ is defined by a set of $N$ nodes $\cV~=~\{1, \ldots, N\}$, edges $\cE~=~\{(i,j)\}$, and a symmetric \textit{adjacency matrix} $\W$. 
$W_{i,j} \in \mathbb{R}^{+}$ is the edge weight if $(i,j) \in \cE$, and $W_{i,j} = 0$ otherwise. Diagonal \textit{degree matrix} $\D$ has diagonal entries $D_{i,i} = \sum_{j} W_{i,j}, \forall i$. 
A \textit{combinatorial graph Laplacian matrix} $\L$ is defined as $\L~\triangleq~\D - \W$ \cite{ortega18ieee}. Further, a \textit{transition matrix} $\A$ is defined as $\A~\triangleq~\D^{-1}\W$~\cite{chen2018}. By definition $\sum_{j\in\mathcal{N}_{i}}A_{i,j}=1$. In general,  a vector $\mathbf{x}~=~[x_{1}\hdots x_{N}]^{\top}\in\mathbb{R}^{V}$ can be interpreted as a graph signal, where $x_i$ is a scalar value assigned to node $i\in \cV$. Further, for a given graph signal $\x$, a weighted average signal value at node $i$ around its neighbors can be computed as
\begin{equation}
    \bar{x}_{i}=(\A\x)_{i}=\sum_{j\in\cN_{i}}A_{i,j}x_{j},
\label{eq:average}    
\end{equation}
where $\cN_{i}$ is the 1-hop neighborhoods of node $i$. Moreover, the second difference of the graph signal $\x$ at node $i$ is given as
\begin{equation}
    \tilde{x}_{i}=(\L\x)_{i}=L_{i,i}x_{i}+\sum_{j\in\cN_{i}}L_{i,j}x_{j}.
\label{eq:sec_diff}    
\end{equation}

\subsubsection{Graph construction for a 3D point cloud}
\label{sec:graph_cons}

To enable graph-based feature-extraction of $n$ 3D points, a neighborhood graph is first constructed. In particular, an undirected positive graph $\mathcal{G} = (\mathcal{V}, \mathcal{E}, \W)$ is created, consisting of a node set $\mathcal{V}$ of size $n$ (where each node represents a 3D point) and an edge set $\mathcal{E}$ defined by $(i,j,W_{i,j})$, where $i \neq j$, $i,j \in \mathcal{V}$, and $W_{i,j} \in \mathbb{R}^{+}$.

Each 3D point (graph node)  is connected to its $k$ nearest neighbors $j$ in Euclidean distance, so that each point can be filtered with its $k$ neighboring points under a graph-structured data kernel \cite{ortega18ieee,cheung2018}.

In the graph-based point cloud processing literature \cite{dinesh2020ICASSP, dinesh2019, dinesh2020ICIP, wei_TSP2020}, using pairwise Euclidean distance $\|\p_i - \p_j\|^2_2$ to compute edge weight $W_{i,j}$ between nodes $i$ and $j$ is popular, \ie,
\begin{equation}
W_{i,j}=
 \exp \left\{-\frac{\norm{\mathbf{p}_{i}-\mathbf{p}_{j}}_{2}^{2}}{\sigma^{2}}\right\}, 
\label{eq:edge_weight}
\end{equation}where $\mathbf{p}_{i}\in \mathbb{R}^{3}$ is the 3D coordinate of point $i$ and  $\sigma$ is a parameter. In numerous graph-based point cloud processing works \cite{dinesh2020ICASSP, dinesh2019, dinesh2020ICIP,Fu2020_ICME, qi2019, zeng2020, Dinesh2023, DineshTIP2022}, $\sigma$ was manually chosen so that edge weight $W_{i,j}$ is large (close to 1) if 3D points $i, j$ are physically close, and small (close to 0) otherwise.

\subsection{Graph-based feature extraction}
Based on these notions, three sets of features are computed for each 3D point in a given point cloud as follows.
\subsubsection{3D point coordinates-based features}
\label{sec:coor} 
A point cloud $\P \in \mathbb{R}^{n\times3}$ is a set of $n$ points in a 3D space, $\P=[\p_{i}], i=1,2, ..., n$, where each point, $\p_{i}=(p_{x,i},p_{y,i},p_{z,i})$, is represented by its x-y-z coordinates. The point cloud can also be represented as $\P~=~[\p_x ; \p_y ; \p_z ]$, where $\p_x$, $\p_y$, and $\p_z$ are the $n$-dimensional column-vectors representing $x$\nobreakdash-, $y$-, and $z$-coordinates. Here, we consider $\p_x$, $\p_y$, and $\p_z$ as three graph signals on graph $\cG$ constructed from a given point cloud as in Section~\ref{sec:graph_cons}. Therefore, using~(\ref{eq:average}), one can easily compute the weighted average 3D coordinate at node $i$ as follows:
\begin{equation}
    \bar{\p}_{i}=\left(\left(\A\P\right)^{\top}\right)(:,i)=\sum_{j\in \mathcal{N}_{i}}A_{i,j}\p_{j},
\label{eq:avg_coord}    
\end{equation}
where $\left(\left(\A\P\right)^{\top}\right)(:,i)$ is the $i$-th column of $\left(\A\P\right)^{\top}$. 
Further, according to~(\ref{eq:sec_diff}), the second difference of 3D coordinates $\P$ at node $i$ can be written as follows:
\begin{equation}
    \widetilde{\p}_{i} = \left(\left(\L\P\right)^{\top}\right)(:,i)=L_{i,i}\p_i+\sum_{j\in\mathcal{N}_{i}}L_{i,j}\p_{j}.
\label{eq:coor_diff}    
\end{equation}
Now, the $x$\nobreakdash-, $y$-, and $z$-coordinates $\bar{\p}_i$ and $\widetilde{\p}_{i}$ are considered as one set of graph-based features at point $i$.

\subsubsection{Local variation-based features}
\label{sec:var}
Similar to~\cite{chen2018}, local variation at point $i$ can be quantified as:
\begin{equation}
    v_i=\norm{\p_i -\bar{\p}_{i}}_{2},
\label{eq:loc_var}    
\end{equation}
where $\bar{\p}_{i}$ is in~(\ref{eq:avg_coord}). Here, one can easily see that $v_{i}$ is the Euclidean distance between point $\p_i$ and the weighted average of its neighbors. Therefore, when the value of $v_i$ is high, the point $p_{i}$ cannot be well approximated from those of its neighboring points, and hence the point $\p_i$ is thus likely to be a point of an edge, corner, or valley. 

Further, we consider $\v=[v_1 \hdots v_n]^{\top}$ as a graph signal of the graph $\cG$ constructed from the given point cloud as discussed in Section~\ref{sec:graph_cons}. Then, according to~(\ref{eq:average}), weighted average signal value at node $i$ is given as:
\begin{equation}
    \bar{v}_{i}=\sum_{j\in \mathcal{N}_{i}}A_{i,j}v_{j}.
\label{eq:avg_var}    
\end{equation}
Using~(\ref{eq:sec_diff}), second difference of signal $\v$ at node $i$ is given as:
\begin{equation}
    \tilde{v}_{i}=L_{i,i}v_i+\sum_{j\in\mathcal{N}_{i}}L_{i,j}v_{j}.
\label{eq:diff_var}    
\end{equation}
Now, we consider $v_i$,  $\bar{v}_i$ and $\tilde{v}_{i}$ are the second set of graph-based features at point $i$.

{
\renewcommand{\arraystretch}{1.4}%
\begin{table*}[t]
\centering
\caption{Fourteen features created for the $i$-th point in a given point cloud.}
\scriptsize
\begin{tabular}{|c|c|}
\hline
Feature symbol & Explanation \\ \hline \hline

     $f_{1}^{i}$ & $v_{i}$. See~(\ref{eq:loc_var}) and Section~\ref{sec:var}.         \\ \hline
      $f_{2}^{i}$         &  $x$ coordinate of $\bar{\p}_{i}$. See~(\ref{eq:avg_coord}) and Section~\ref{sec:coor}.        \\ \hline
   $f_{3}^{i}$         &  $y$ coordinate of $\bar{\p}_{i}$. See~(\ref{eq:avg_coord}) and Section~\ref{sec:coor}.        \\ \hline
  $f_{4}^{i}$         &  $z$ coordinate of $\bar{\p}_{i}$. See~(\ref{eq:avg_coord}) and Section~\ref{sec:coor}.        \\ \hline
   $f_{5}^{i}$         &  $x$ coordinate of $\widetilde{\p}_{i}$. See~(\ref{eq:coor_diff}) and Section~\ref{sec:coor}.        \\ \hline
   $f_{6}^{i}$         &  $y$ coordinate of $\widetilde{\p}_{i}$. See~(\ref{eq:coor_diff}) and Section~\ref{sec:coor}.        \\ \hline
   $f_{7}^{i}$         &  $z$ coordinate of $\widetilde{\p}_{i}$. See~(\ref{eq:coor_diff}) and Section~\ref{sec:coor}.        \\ \hline
    $f_{8}^{i}$           & $\bar{v}_{i}$. See~(\ref{eq:avg_var}) and Section~\ref{sec:var}.         \\ \hline
     $f_{9}^{i}$          & $\tilde{v}_{i}$. See~(\ref{eq:diff_var}) and Section~\ref{sec:var}.          \\ \hline
      $f_{10}^{i}$       & Euclidean distance between point $i$ and the centroid of the point cloud.          \\ \hline
      $f_{11}^{i}$         & The number of points inside a ball of radius $r$ and center $\p_{i}$.         \\ \hline
       $f_{12}^{i}$        & $h_{i}$. See Section~\ref{sec:LPF}.         \\ \hline
        $f_{13}^{i}$       & $\bar{h}_{i}$. See Section~\ref{sec:LPF}.         \\ \hline
         $f_{14}^{i}$      & $\tilde{h}_{i}$. See Section~\ref{sec:LPF}.         \\ \hline
\end{tabular}
\label{tab:features}
\end{table*}
}

\subsubsection{Low-pass filter-based features}
\label{sec:LPF}
For graph signals $\p_{x}$, $\p_{y}$, $\p_{z}$ with respect to the graph $\cG$, the corresponding \textit{low-pass filter} (LPF) signals $\q_{x}^{*}$, $\q_{y}^{*}$, $\q_{z}^{*}$ can be obtained by minimizing the following optimization problem (See~\cite{schoenenberger2015} for details):
\begin{equation}
    \q^{*}=\arg\min_{\q}\norm{\p-\q}_{2}^{2}+\gamma\left(\q_{x}^{\top}\L\q_{x}+\q_{y}^{\top}\L\q_{y}+\q_{z}^{\top}\L\q_{z}\right),
\label{eq:LPF_formulation}    
\end{equation}
where $\q^{*}~=~[\left(\q_x^{*}\right)^{\top} \hspace{4pt} \left(\q_y^{*}\right)^{\top}\hspace{4pt} \left(\q_z^{*}\right)^{\top} ]^{\top}$, $\q~=~[\q_x^{\top} \hspace{4pt} \q_y^{\top}\hspace{4pt} \q_z^{\top} ]^{\top}$, $\p~=~[\p_x^{\top} \hspace{4pt} \p_y^{\top}\hspace{4pt} \p_z^{\top} ]^{\top}$, and $\gamma>0$ is a regularization parameter. Since~(\ref{eq:LPF_formulation}) is a \textit{quadratic programming} (QP) problem, its solution can be obtained by solving the following system of linear equations:
\begin{equation}
    \left(\I+\gamma\bar{\L}\right)\q^{*}=\p,
\label{eq:LPF_sol}    
\end{equation}
where $\bar{\L}=\text{diag}\{\L, \L, \L\}$. Since $\L$ is a \textit{positive semi-definite} (PSD) matrix by definition~\cite{chung1997}, one can easily see that $\bar{\L}$ is also a PSD matrix. Hence $(\I+\gamma\bar{\L})$ is a \textit{positive definite} (PD) matrix for $\gamma>0$ and thus invertible. 

Moreover, we define another graph signal $\h=[h_1 \hdots h_n]$, where $h_i=\norm{\p_i-\q_i^{*}}_2$. Here, $\q_i^{*}\in \mathbb{R}^{3}$ is $x$, $y$, $z$ coordinates of the LPF coordinates at point $i$ obtained from~(\ref{eq:LPF_sol}). Further, similar to~(\ref{eq:avg_var}) and~(\ref{eq:diff_var}), we can compute the weighted average signal value at node $i$ and the second difference of $\h$ at node $i$ (denoted as $\bar{h}_{i}$ and $\tilde{h}_{i}$, respectively.) Then, we consider $h_i$, $\bar{h}_i$ and $\tilde{h}_{i}$ are the third set of graph-based features at point $i$. 

In addition to the aforementioned graph-based features, the following two features are computed for each point $i$:
\begin{enumerate}
    \item Euclidean distance between point $i$ and the centroid of the corresponding point cloud.
    \item The number of points inside a ball of radius $r$ and center $\p_i$. In this paper, we manually choose $r=0.1$.
\end{enumerate}

Altogether, fourteen features (denoted as $\left\{f^i_j\,|\,j=1,...,14\right\}$) have been created for any given point $i$, as summarized in Table~\ref{tab:features}.Table~\ref{tab:features}. These are: the edge intensity value ($f^i_1$), weighted average of 3D coordinates around neighboring points ($f^i_2$, $f^i_3$, $f^i_4$), second difference of 3D coordinates ($f^i_5$, $f^i_6$, $f^i_7$), the weighted average of edge intensity values around neighboring points ($f^i_8$), the second difference of edge intensity values ($f^i_9$), the distance from the centroid ($f^i_{10}$), the number of points inside a ball of radius $r=0.1$ ($f^i_{11}$), the distance between actual 3D points and the low-pass-filtered (LPF) 3D points ($f^i_{12}$), the weighted average of LPF distance around neighbors ($f^i_{13}$) and the second difference of LPF distance ($f^i_{14}$).

\section{Feature Analysis and New Attack}
\label{sec:Proposed_Method}
In this section, a methodology is presented to examine how well the features introduced in Section~\ref{sec:features} are able to predict adversarial drop points.  
First, each point is assigned an \emph{adversarial score} 
that reflects how much it contributes to the corresponding model's prediction.
Then, multiple linear regression is used to examine how predictive are various features of the adversarial score. 

\subsection{Adversarial score}
Let $\digamma:\mathbb{R}^{n\times3} \rightarrow \{1,2,...,C\}$ 
be a trained $C$-class classifier, which maps an input point cloud $\P$ to a class $c_{t}\in\{1,2,...,C\}$, 
such that $\digamma(\P) = c_{t}$. 
An adversarial attack aims to deceive the classifier by changing the point cloud $\P$ to $\P^{adv}$ so that $\digamma(\P^{adv}) \neq c_{t}$, while usually also requiring $\P^{adv} \approx \P$. 

Saliency score~(\ref{eq:saliency_score}) is an indicator of the sensitivity of the classifier to a perturbation of an input point. It has been shown~\cite{zheng2019pointcloud} to be effective in determining adversarial points, in the sense that perturbing or removing points with high saliency score can create adversarial examples. One issue with the saliency score is that its dynamic range can be variable. Hence, the adversarial score is defined by normalizing the saliency score to the range $[0, 1]$. Specifically, for the $i$-th point $\p_i$, the adversarial score is defined as
\begin{equation}
    z(\p_i)=\frac{s_i-\min\{s_1,...,s_n\}}{\max\{s_1,...,s_n\} - \min\{s_1,...,s_n\}},
    \label{eq:adversarial_score}
\end{equation}
where $s_i$ is given in~(\ref{eq:saliency_score}). In the next section, we examine how well our fourteen features are able to predict $z(\p_i)$.

\subsection{Multiple linear regression}
\label{sec:MLR}
Multiple linear regression analysis~\cite{kutner_applied_2005,eberly2007multiple} is employed to examine how predictive are the features defined in Section~\ref{sec:features} of the adversarial score $z(\p_i)$ in~(\ref{eq:adversarial_score}). Note that this idea may seem strange to start with: $z(\p_i)$ depends on the classifier model, via the loss $\mathcal{L}$ in the saliency score $s_i$ in~(\ref{eq:saliency_score}), while the features in Section~\ref{sec:features} do not! Yet, as the results will show, some of our features are fairly predictive of $z(\p_i)$.

\begin{figure*}
    \centering
    \includegraphics[width=0.9\textwidth]{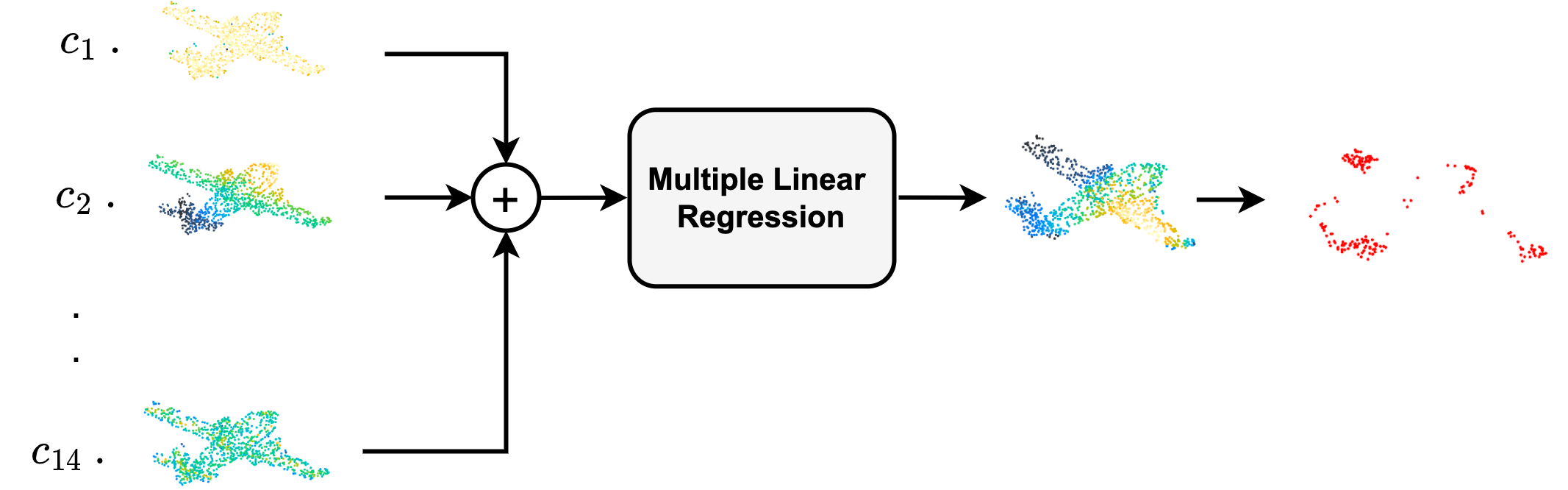}
     \caption{An overview of multiple linear regression analysis. Fourteen features are computed for each point in the point cloud. In the figure, feature values are represented with different colors, with black corresponding to high values and light yellow corresponding to low values. At each point, features are linearly combined using the coefficients estimated on the training data, whose significance is determined using statistical testing. In the final experiment, the $N$ points with the highest predicted score are compared against the $N$ points with the highest true adversarial score.}
    \label{fig_proposed}
\end{figure*}

For a given point $\p_i$, its adversarial score $z(\p_i)$ and its features $\left\{f^i_j\,|\,j=1,...,14\right\}$, we set up the following multiple linear regression model:
\begin{equation}
    z(\p_i) \approx \sum_{j=1}^{14}c_j\cdot f^i_j,
    \label{eq:MLR}
\end{equation}
where $c_j$ are the regression coefficients. An illustration is shown in Fig.~\ref{fig_proposed}. The coefficients will be fitted on a set of points from a dataset of point clouds. Naturally, there is a variation in the coefficient values across different point clouds. To determine whether a particular feature is predictive of $z(\p_i)$, we run the following hypothesis test for each coefficient $c_j$ as a t-test~\cite{kutner_applied_2005}:
\begin{equation}
    \begin{aligned}
        H_0: c_j = 0, \\
        H_1: c_j \neq 0. \\
    \end{aligned}
    \label{eq:hypothesis_test}
\end{equation}
Here, the null hypothesis $H_0$ is $c_j=0$. If the null hypothesis cannot be rejected for a particular coefficient $c_j$, it means that, given the variation, the coefficient is not significantly different from zero. The interpretation would be that the corresponding feature $f^i_j$ does not contribute significantly to the prediction of the adversarial score $z(\p_i)$. On the other hand, for coefficients $c_j$ where the null hypothesis can be rejected, we can conclude that the corresponding feature $f^i_j$ is statistically significantly predictive of $z(\p_i)$. As will be seen in Section~\ref{sec:results}, some features are predictive of $z(\p_i)$, others are not.

\subsection{New drop attack}
\label{sec:no-box_attack}
Let $\mathcal{F}\subseteq \{1,2,...,14\}$ be the set of indices of features that have been determined to have a statistically significant effect on the adversarial score, as described in Section~\ref{sec:MLR}. For each point $\p_i$ in a given point cloud $\P$, we compute the predicted adversarial score using the statistically significant features as
\begin{equation}
    \mathfrak{z}(\p_i) = \sum_{j\in \mathcal{F}}c_j\cdot f^i_j,
    \label{eq:predicted_score}
\end{equation}
where $c_j$ is the coefficient of the $j$-th feature estimated through multiple linear regression analysis. Note that $z(\p_i)$ is the true adversarial score obtained from~(\ref{eq:saliency_score}) and~(\ref{eq:adversarial_score}), which requires access to the target DNN model, while $\mathfrak{z}(\p_i)$ is the predicted adversarial score computed from the point cloud itself. Once the predicted adversarial scores $\mathfrak{z}(\p_i)$ are computed for all points in the point cloud, the points are sorted in decreasing order of $\mathfrak{z}(\p_i)$. To create a drop-$N$ attack, the top $N$ points on the sorted list are removed from the point cloud. 

\section{Experiments}
\label{sec:results}
\subsection{Experimental setting} 
 The aligned benchmark ModelNet40\footnote{\url{https://github.com/kimianoorbakhsh/LPF-Defense/tree/main/model/Data}} dataset, containing 40 object classes, was used. The dataset employed in this study consists of 9,843 training and 2,468 test point clouds.
 Each point cloud contains 1,024 points. For start, three models -- PointNet, PointNet++, and DGCNN, with implementation from~\cite{wu2020if} -- were used as deep classifiers on the ModelNet40 dataset. Subsequently, another model, PointConv~\cite{wu2019pointconv}, was used to test attack transferability. The approach from~\cite{zheng2019pointcloud} was used to compute saliency scores~(\ref{eq:saliency_score}) iteratively. Specifically,~\cite{zheng2019pointcloud} was used to compute the saliency scores for the initial point cloud, then 10 points with the highest score were removed, then this was repeated for the remaining points until a total of $N$ points were identified. 
Once the $N$ points with the highest saliency score were identified, their adversarial score $z(\p_i)$ was computed as in~(\ref{eq:adversarial_score}).
All experiments were conducted using PyTorch on a machine with a NVIDIA Tesla P100-PCIe card with 16 GB of memory.


\begin{figure}
  \centering
  \begin{tabular}{ c @{\hspace{20pt}} c }
    \includegraphics[width=0.2\textwidth]{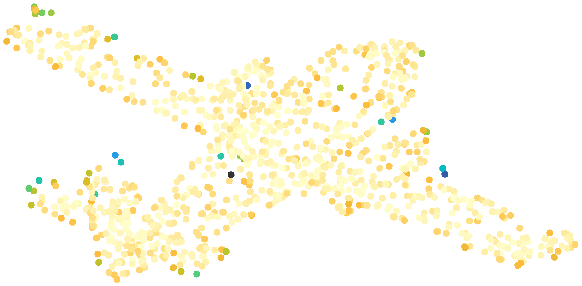} &
      \includegraphics[width=0.2\textwidth]{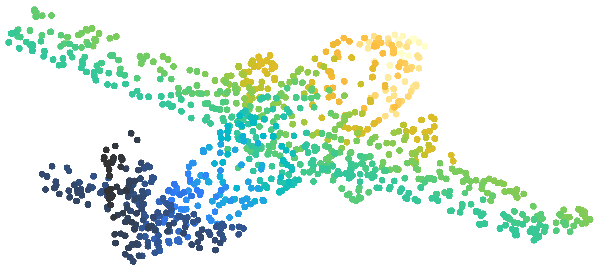} \\
    \small (a) $f_1$ &
      \small (b) $f_2$ \\

    \includegraphics[width=0.2\textwidth]{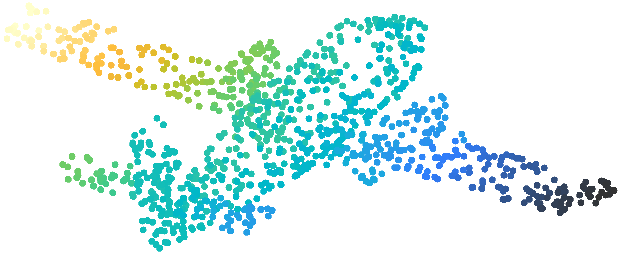} &
      \includegraphics[width=0.2\textwidth]{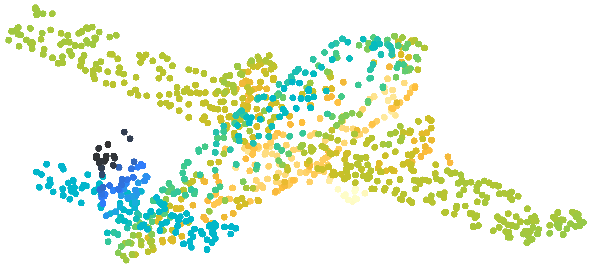} \\
    \small (c) $f_3$ &
      \small (d) $f_4$ \\

    \includegraphics[width=0.2\textwidth]{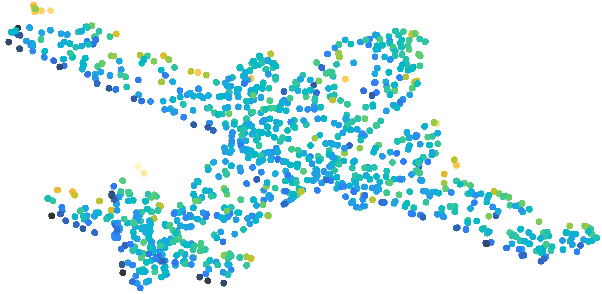} &
      \includegraphics[width=0.2\textwidth]{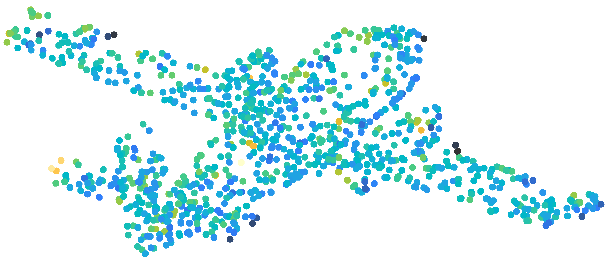} \\
    \small (e) $f_5$ &
      \small (f) $f_6$ \\

    \includegraphics[width=0.2\textwidth]{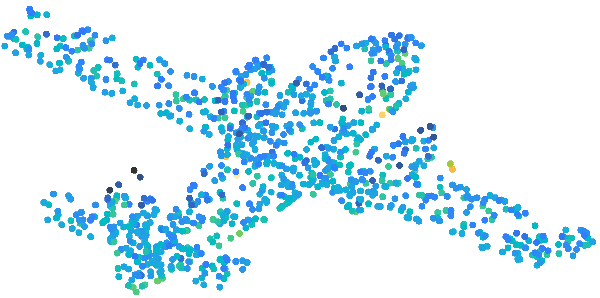} &
      \includegraphics[width=0.2\textwidth]{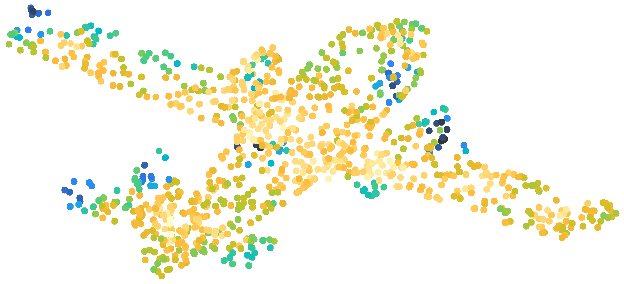} \\
    \small (g) $f_7$ &
      \small (h) $f_8$ \\

    \includegraphics[width=0.2\textwidth]{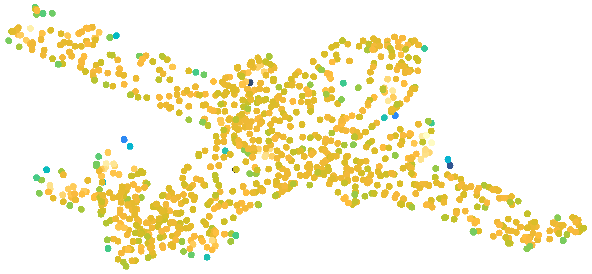} &
      \includegraphics[width=0.2\textwidth]{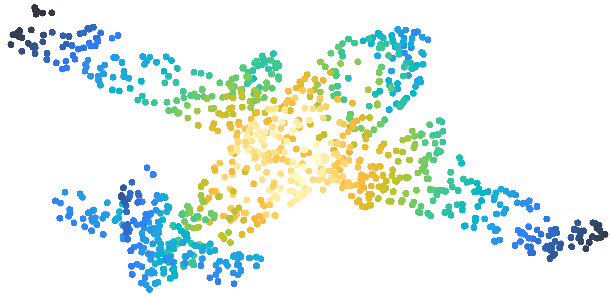} \\
    \small (i) $f_9$ &
      \small (j) $f_{10}$ \\

    \includegraphics[width=0.2\textwidth]{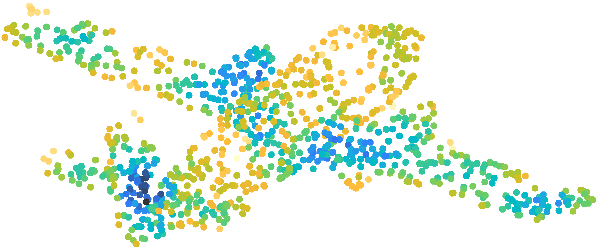} &
      \includegraphics[width=0.2\textwidth]{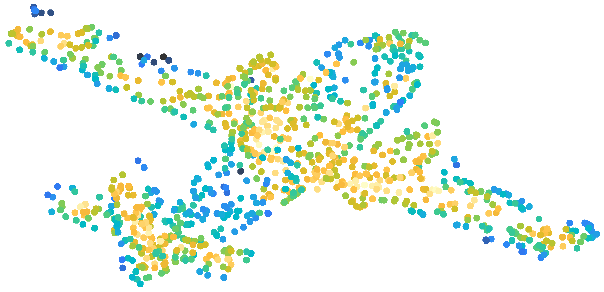} \\
    \small (k) $f_{11}$ &
      \small (l) $f_{12}$ \\

    \includegraphics[width=0.2\textwidth]{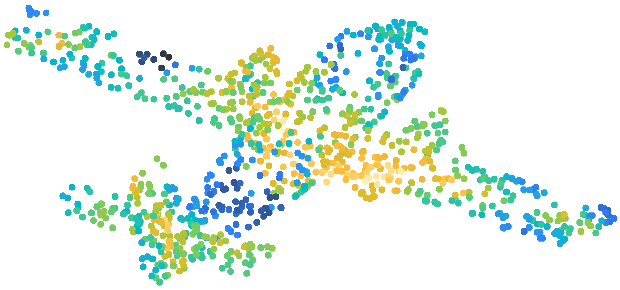} &
      \includegraphics[width=0.2\textwidth]{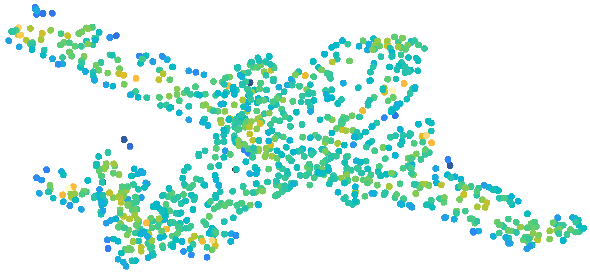} \\
    \small (m) $f_{13}$ &
      \small (n) $f_{14}$ \\
      \\
    \multicolumn{2}{c}{\includegraphics[width=0.35\textwidth]{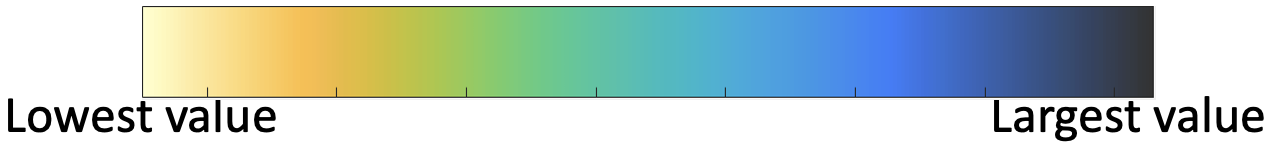}} \\
  \end{tabular}
\caption{Visualization of Fourteen features. Points are colorized by the feature value at each point, according to the shown color map.}
\label{14_Features}
\end{figure}

\subsection{Multiple linear regression analysis} 

Fig.~\ref{14_Features} shows a visualization of the fourteen features defined in Section~\ref{sec:features} on the airplane object. Each point's color reflects the value of the corresponding feature at that point, with  
dark blue indicating high values and light yellow indicating low values.

\begin{figure}
  \centering
  \begin{tabular}{ c @{\hspace{14pt}} c }
    \includegraphics[width=0.23\textwidth]{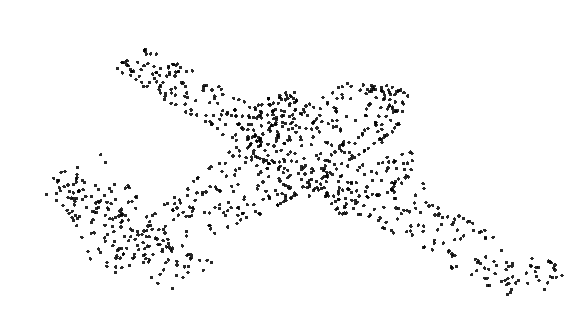} &
      \includegraphics[width=0.23\textwidth]{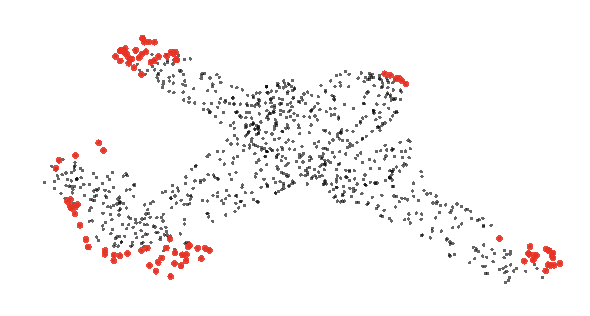} \\
    \small (a) Original &
      \small (b) PointNet \\

    \includegraphics[width=0.23\textwidth]{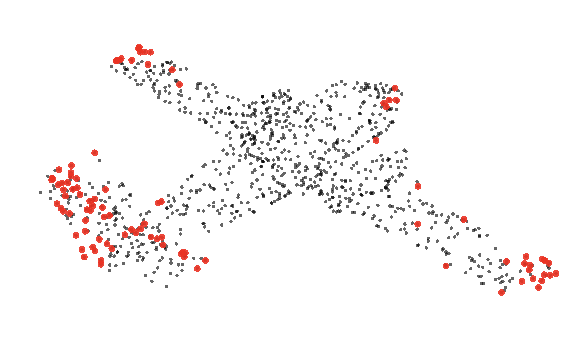} &
      \includegraphics[width=0.23\textwidth]{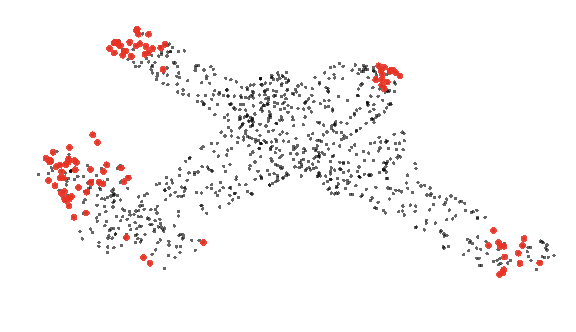} \\
    \small (c) PointNet++ &
      \small (d) DGCNN \\

  \end{tabular}
\caption{Illustration of adversarial points on the airplane object (a); sub-figures (b), (c), and (d) show the 100 points with the highest adversarial score obtained using the corresponding network.}
\label{fig_adv_points}
\end{figure}

Saliency scores $s_i$ in~(\ref{eq:saliency_score}), and therefore also the adversarial scores $z(\p_i)$ in~(\ref{eq:adversarial_score}), tend to be most reliable when $s_i$ is high. Therefore, multiple linear regression analysis was applied only to the $N$ points with the highest scores, where $N\in\{50, 100, 150, 200\}$. 
Fig.~\ref{fig_adv_points} shows a visualization of the 100 points with the highest adversarial score, computed from saliency scores of three models: PointNet, PointNet++, and DGCNN. 

An interesting observation is that, while each network is different, adversarial points tend to cluster in specific regions like the tail, wings, and tips. Comparing Figs.~\ref{14_Features} and~\ref{fig_adv_points}, reveals that some features might be indicative of the adversarial points.

To formally test this idea, multiple linear regression analysis is used, specifically the \texttt{scikit-learn}\footnote{\url{https://scikit-learn.org/stable/}} implementation. For each object in the training set, $N$ points with the highest adversarial score $z(\p_i)$ are selected, and then the regression model~(\ref{eq:MLR}) is fitted."

Hypothesis test~(\ref{eq:hypothesis_test}) is run for each coefficient $c_j$ to see whether the null hypothesis can be rejected at the significance level of $\alpha=0.05$~\cite{kutner_applied_2005}. Those coefficients for which the null hypothesis can be rejected are deemed significantly different from zero, and the corresponding feature is regarded as sufficiently explanatory for $z(\p_i)$. Other coefficients, for which the null hypothesis cannot be rejected at the significance level of $\alpha=0.05$, are deemed insignificant, and the corresponding features regarded as not sufficiently explanatory for $z(\p_i)$. This procedure is repeated for $N\in\{50, 100, 150, 200\}$. The results are shown in Tables~\ref{tab_pointnet},~\ref{tab_pointnet++}, and~\ref{tab_dgcnn} for PointNet, PointNet++, and DGCNN, respectively, where insignificant coefficients are shown as $0$.


\begin{table*}
    \centering
    \caption{Multiple linear regression analyses for $N\in\{50, 100, 150, 200\}$ points with the highest adversarial score derived from PointNet. Significant coefficients (at $\alpha=0.05$) are shown with 3 decimal points precision, while insignificant coefficients are shown as 0. 
    }
    \tiny
    \begin{tabular}{ |c|c|c|c|c|p{0.07cm}|p{0.07cm}|p{0.07cm}|p{0.07cm}|c|c|c|c|c|c|c| } 
        \hline
        
         $N$ & $c_1$ & $c_{2}$ & $c_3$ & $c_4$ & $c_5$ & $c_{6}$ & $c_{7}$ & $c_{8}$ & $c_{9}$ & $c_{10}$ & $c_{11}$ & $c_{12}$ & $c_{13}$ & $c_{14}$ & $R^2$ (\%) \\
        \hline
        \multirow{1}{3em}{50} & -38.043 & 0.007 & 0.005 & -0.009 & 0 & 0 & 0 & 0 & 4.659 & 0.648 & 0.011 & -3.554 & 12.309 & 0.196 & 94.3\% \\ 

        \multirow{1}{3em}{100} & -44.032 & 0.007 & 0.006 & -0.008 & 0 & 0 & 0 & 0 & 5.113 & 0.636 & 0.011 & -3.139 & 11.733 & 0.164 & 94.2\% \\

        \multirow{1}{3em}{150} & -42.295 & 0.007 & 0.006 & -0.007 & 0 & 0 & 0 & 0 & 4.904 & 0.623 & 0.010 & -3.055 & 11.470 & 0.160 & 94.1\% \\     
        
        \multirow{1}{3em}{200} & -41.451 & 0.007 & 0.005 & -0.007 & 0 & 0 & 0 & 0 & 4.819 & 0.611 & 0.010 & -2.969 & 11.207 & 0.153 & 93.9\% \\  
        \hline
    \end{tabular}
    \label{tab_pointnet}
\end{table*}


\begin{table*}
    \centering
    \caption{Multiple linear regression analyses for $N\in\{50, 100, 150, 200\}$ points with the highest adversarial score derived from PointNet++. Significant coefficients (at $\alpha=0.05$) are shown with 3 decimal points precision, while insignificant coefficients are shown as 0. 
    }
    
    \tiny
    \begin{tabular}{ |c|c|c|c|c|p{0.07cm}|p{0.07cm}|p{0.07cm}|p{0.07cm}|c|c|c|c|c|c|c| } 
        \hline
         $N$ & $c_1$ & $c_{2}$ & $c_3$ & $c_4$ & $c_5$ & $c_{6}$ & $c_{7}$ & $c_{8}$ & $c_{9}$ & $c_{10}$ & $c_{11}$ & $c_{12}$ & $c_{13}$ & $c_{14}$ & $R^2$ (\%) \\
        \hline
        
         \multirow{1}{3em}{50} & -54.854 & 0.009 & 0.006 & -0.007 & 0 & 0 & 0 & 8.859 & 6.112 & 0.649 & 0.011 & -2.665 & 11.375 & 0.122 & 94.3\% \\ 
        
        \multirow{1}{3em}{100} & -49.452 & 0.008 & 0.005 & -0.007 & 0 & 0 & 0 & 6.851 & 5.544 & 0.636 & 0.010 & -2.908 & 11.370 & 0.148 & 94.2\% \\ 

        \multirow{1}{3em}{150} & -44.927 & 0.008 & -0.006 & -0.008 & 0 & 0 & 0 & 3.120 & 5.125 & 0.624 & 0.010 & -3.067 & 11.320 & 0.161 & 94.1\% \\ 

        \multirow{1}{3em}{200} & -43.109 & 0.007 & 0.005 & -0.007 & 0 & 0 & 0 & 2.489 & 4.938 & 0.612 & 0.010 & -3.057 & 11.091 & 0.163 & 93.9\% \\         
       
        \hline
    \end{tabular}
    \label{tab_pointnet++}
\end{table*}

\begin{table*}
    \centering
    \caption{Multiple linear regression analyses for $N\in\{50, 100, 150, 200\}$ points with the highest adversarial score derived from DGCNN. Significant coefficients (at $\alpha=0.05$) are shown with a precision of 3 decimal points, while insignificant coefficients are shown as 0. 
    }
    \tiny
    \begin{tabular}{ |c|c|c|c|c|p{0.07cm}|p{0.07cm}|p{0.07cm}|p{0.07cm}|c|c|c|c|c|c|c| } 
        \hline
        $N$ & $c_1$ & $c_{2}$ & $c_3$ & $c_4$ & $c_5$ & $c_{6}$ & $c_{7}$ & $c_{8}$ & $c_{9}$ & $c_{10}$ & $c_{11}$ & $c_{12}$ & $c_{13}$ & $c_{14}$ & $R^2$ (\%) \\
        \hline
        
         \multirow{1}{3em}{50} & -52.555 & 0.006 & 0.006 & -0.008 & 0 & 0 & 0 & 10.169 & 5.870 & 0.648 & 0.011 & -3.058 & 11.745 & 0.157 & 94.4\% \\ 
        
        \multirow{1}{3em}{100} & -46.105 & 0.006 & 0.005 & -0.007 & 0 & 0 & 0 & 4.473 & 5.241 & 0.636 & 0.011 & -3.257 & 11.818 & 0.177 & 94.3\% \\ 

        \multirow{1}{3em}{150} & -43.374 & 0.007 & 0.005 & -0.007 & 0 & 0 & 0 & 3.352 & 4.960 & 0.623 & 0.011 & -3.179 & 11.540 & 0.173 & 94.2\% \\ 

        \multirow{1}{3em}{200} & -41.795 & 0.006 & 0.004 & -0.007 & 0 & 0 & 0 & 3.585 & 4.806 & 0.610 & 0.011 & -3.153 & 11.330 & 0.174 & 94.0\% \\    
                
        \hline
    \end{tabular}
    \label{tab_dgcnn}
\end{table*}

Consider Table~\ref{tab_pointnet} first. 
Four coefficients are insignificant: $c_5$, $c_6$, $c_7$, and $c_8$, and these are shown as $0$, while the others are significant at $\alpha=0.05$. The last column contains the $R^2$ coefficient of determination~\cite{kutner_applied_2005}, shown as a percentage. $R^2$ is around 94\%, which is fairly high and indicates good agreement between the data and the model

As the number of points $N$ increases, it becomes harder for the model to fit the data, hence the drop in $R^2$. 
Similar behavior can be seen in Tables~\ref{tab_pointnet++} and~\ref{tab_dgcnn} as well, although in these cases, 
$c_8$ is significant. 

Another observation is that there is a large overlap between the set of significant coefficients for PointNet, PointNet++, and DGCNN, even though these DNN models are quite different. This suggests that certain features of the point cloud itself may be able to predict adversarial points for different DNN models, and this is what makes new attacks possible. Specifically, we see that edge intensity ($f_1$), weighted coordinate-based features ($f_2$, $f_3$, $f_4$), local variation-based feature ($f_9$), distance from the centroid ($f_{10}$), and LPF-based features ($f_{11}$, $f_{12}$, $f_{13}$, $f_{14}$) are all indicative of the adversarial score for all three DNN models.  

Tables~\ref{tab_pointnet},~\ref{tab_pointnet++}, and~\ref{tab_dgcnn} 
also give the coefficients of the linear combination of features that best predicts the adversarial score for each case. 
For example, if we were to predict the adversarial score derived from PointNet for the top $N=150$ points, then based on the corresponding row in Table~\ref{tab_pointnet}, we could use the following linear model: 
\begin{equation}
    \begin{split}
        \mathfrak{z}(\p_i) = &- 42.295\cdot f^i_1 + 0.007\cdot f^i_{2} + 0.006\cdot f^i_{3}\\ 
        &- 0.007\cdot f^i_{4} + 4.904\cdot f^i_{9} + 0.623\cdot f^i_{10}\\
        & + 0.010\cdot f^i_{11} - 3.055\cdot f^i_{12} + 11.470\cdot  f^i_{13}\\ 
        & + 0.160\cdot f^i_{14}.
    \end{split}
\end{equation}

Although the best coefficients are different for different cases (i.e., different $N$'s and tables), it is noticeable that the coefficient values do not change too much - they are the same order of magnitude across all cases.  It can also be observed from  Tables~\ref{tab_pointnet},~\ref{tab_pointnet++}, and~\ref{tab_dgcnn} that the regression coefficients 
do not vary greatly across the different cases, regardless of whether the linear model was fit to the adversarial scores from  
PointNet, PointNet++, or DGCNN. This consistency in the coefficient values suggests that the features identified as significant by the regression analysis are robust predictors of adversarial points, regardless of the DNN model. 
We test this observation in subsequent experiments in Sections~\ref{sec:success_rate} and~\ref{sec:transferability}.

\begin{figure}
  \centering
  \begin{tabular}{ c @{\hspace{14pt}} c }
    \includegraphics[width=0.49\textwidth]{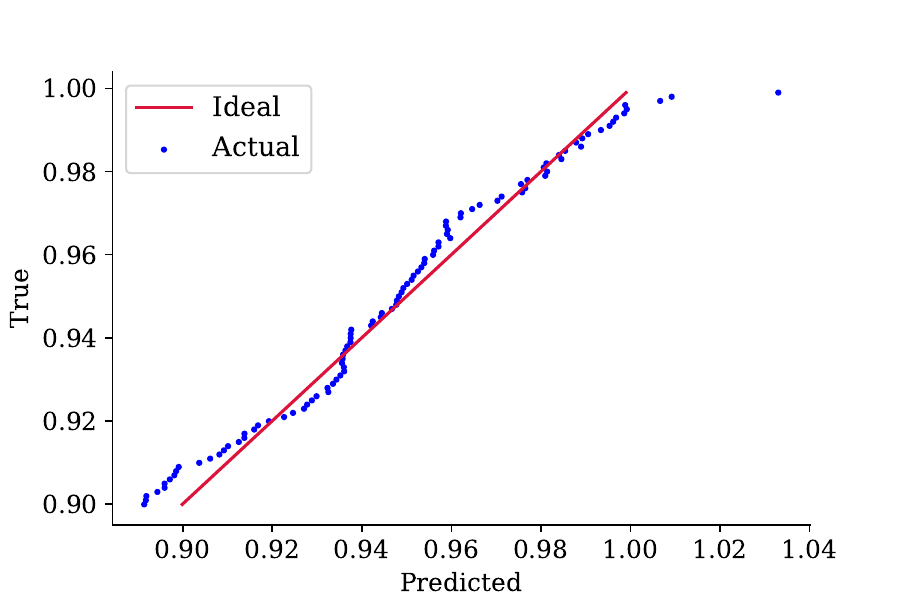} &
      \includegraphics[width=0.49\textwidth]{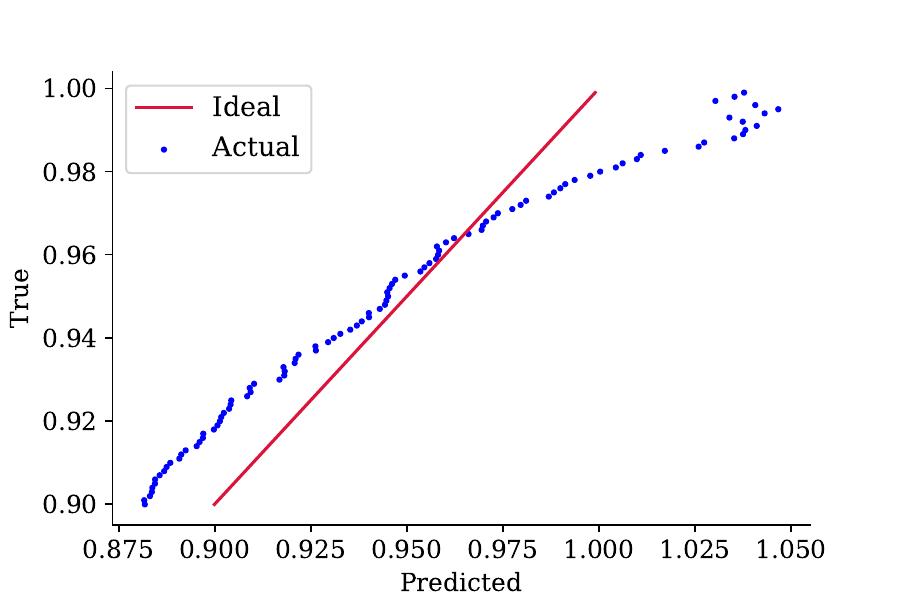} \\

    \includegraphics[width=0.49\textwidth]{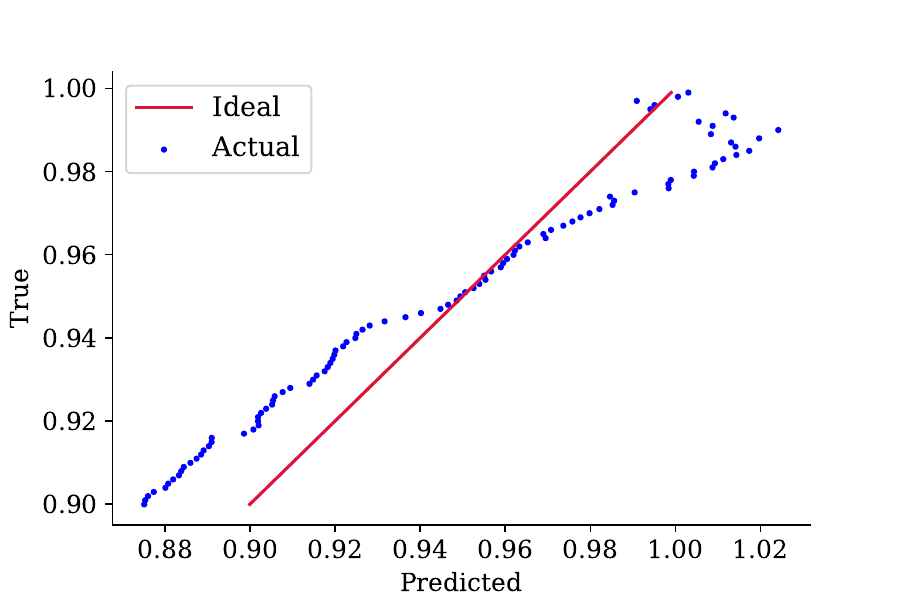} &
      \includegraphics[width=0.49\textwidth]{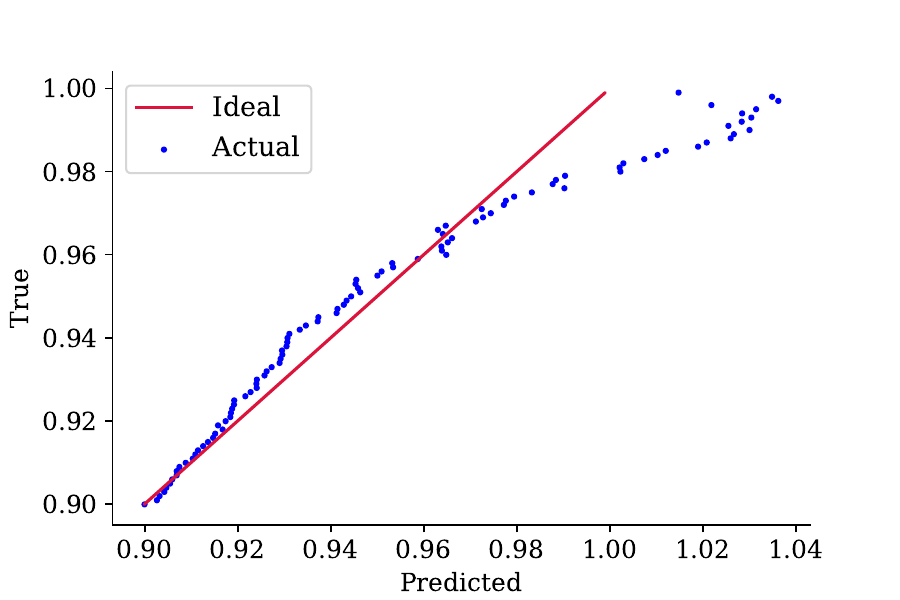} \\

  \end{tabular}
\caption{Predicted vs. true adversarial scores for four point clouds. A perfect prediction would correspond to the straight red line, actual predictions give blue scatter plots.}
\label{fig_regression_error}
\end{figure}

To examine how accurately the regression model predicts the true adversarial scores, we show four graphs in Fig.~\ref{fig_regression_error}. These graphs correspond to four point clouds whose adversarial scores were computed based on PointNet. The blue scatter plots show $N=100$ points as pairs of values (\texttt{Predicted}, \texttt{True}), where \texttt{Predicted} is the adversarial score $\mathfrak{z}(\p_i)$ predicted by the linear model~(\ref{eq:MLR}), while \texttt{True} is the true adversarial score $z(\p_i)$ from~(\ref{eq:adversarial_score}). Ideally, if the predictions were perfect, all the blue points would fall on the straight red line. Although this is not the case in practice, the results are close and, more importantly, the rank order is mostly preserved. This means that to find the $N$ points with the highest true adversarial score, many can be identified by computing the predicted score from the features and taking the $N$ points with the highest predicted score.

\begin{figure*}
  \centering
  \begin{tabular}{ c @{\hspace{5pt}} c@{\hspace{5pt}} c }
    \includegraphics[width=0.32\textwidth]{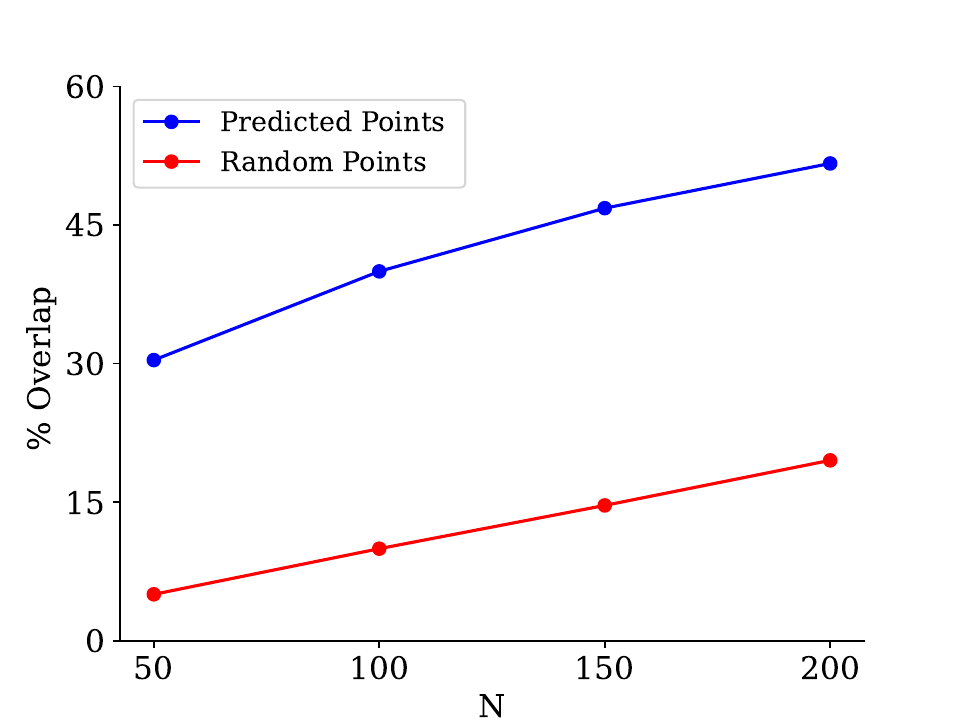} &
      \includegraphics[width=0.32\textwidth]{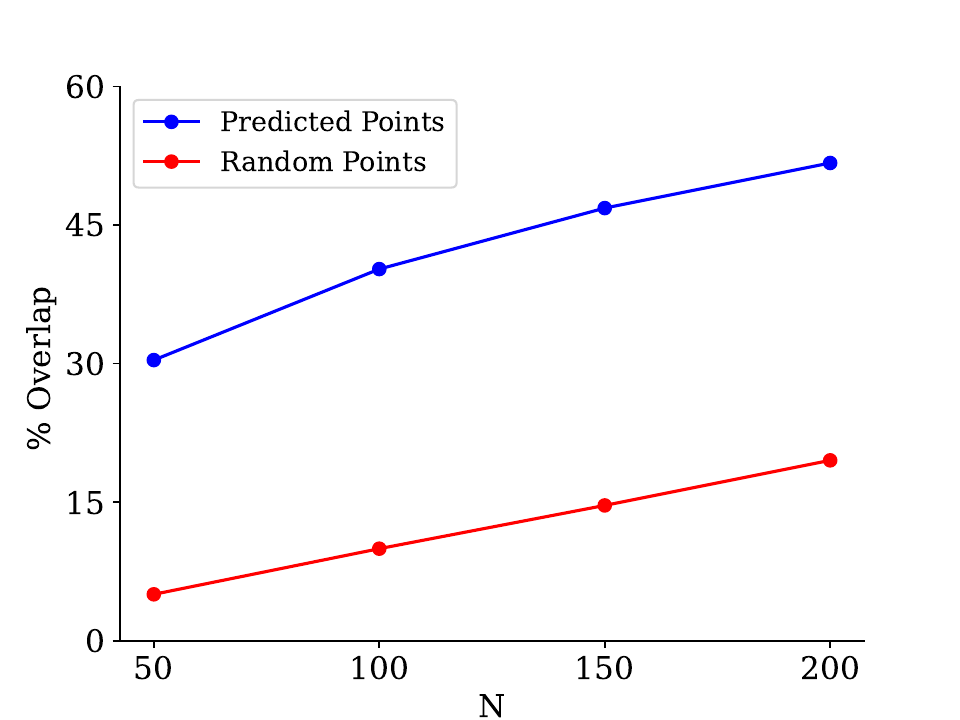} &
      \includegraphics[width=0.32\textwidth]{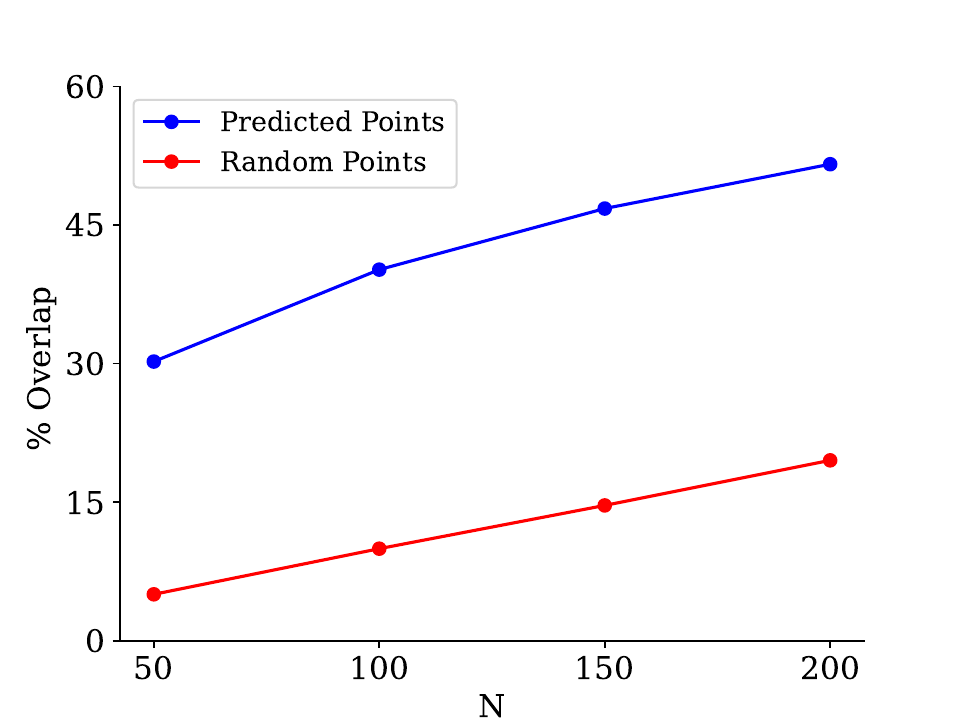} \\
    \small (a) PointNet &
    \small (b) PointNet++ &
    \small (c) DGCNN \\


  \end{tabular}
\caption{Percentage overlap between top-$N$ points with predicted and true adversarial scores generated by (a) PointNet, (b) PointNet++, and (c) DGCNN. Overlap with randomly chosen points is also shown.}
\label{fig_comparison}
\end{figure*}


To test this idea further, we compute the overlap between the set of top-$N$ points according to the true adversarial score $z(\p_i)$ from~(\ref{eq:adversarial_score}) and the set of top-$N$ points according to the predicted adversarial score $\mathfrak{z}(\p_i)$ from~(\ref{eq:MLR}). The results are shown in Fig.~\ref{fig_comparison}. The three graphs correspond to the three networks (PointNet, PointNet++, and DGCNN) based on which the true adversarial scores are computed. Each graph shows the percentage overlap for $N\in\{50, 100, 150, 200\}$, for the prediction based on proposed features, as well as a random selection of points. The gap between the two lines indicates how much better is the prediction based on the proposed features compared to a random guess. As seen in the graphs, the prediction based on the proposed features is about 25-30\% better than a random guess and achieves over 50\% overlap with the true adversarial drop points for $N=200$.

\subsection{Attack success rate}
\label{sec:success_rate}
The attack success rate measures how effective a particular adversarial attack is at fooling the target deep model. The attack success rate is typically reported as a percentage of cases in which the target model was fooled on a given dataset, with a higher number indicating a more successful attack. The next experiment examines the attack success rates on the ModelNet40 test set.

Since there are no existing attacks based on intrinsic characteristics of the point clouds in 3D point cloud classification, we pick as our ``Baseline'' method an attack based on saliency score~(\ref{eq:saliency_score}), which is the un-normalized version of the true adversarial score~(\ref{eq:adversarial_score}). The ``Baseline'' attack involves dropping 100 or 200 points with the highest true saliency scores. This is a white-box attack and requires access to the target DNN model so that gradients relative to the loss function can be computed in~(\ref{eq:saliency_score}). Although the white-box attack clearly has an advantage over the proposed attack, the comparison still helps us gain insight into how much the attack success rate can be improved by having direct access to the target DNN model. The second method (``Proposed'') is the proposed attack, which drops 100 or 200 points with the highest predicted adversarial score $\mathfrak{z}(\p_i)$ from~(\ref{eq:predicted_score}), using the \emph{averaged} significant coefficients from Tables~\ref{tab_pointnet},~\ref{tab_pointnet++}, and~\ref{tab_dgcnn}. The purpose of using the average coefficients from the three tables is to avoid having the prediction model~(\ref{eq:predicted_score}) be fitted to any particular target DNN. The third method (``Random'') drops 100 or 200 randomly selected points.

Table~\ref{tab:Attack success rate} shows the attack success rate results. Note that the table also includes results for PointConv~\cite{wu2019pointconv}, a model that was not included in the computation of coefficients in Tables~\ref{tab_pointnet},~\ref{tab_pointnet++}, and~\ref{tab_dgcnn}. As expected, the Baseline method achieves the highest success rate, since it is a white-box attack that uses a particular target DNN model to figure out which points are the most important for that model. 

However, the Proposed attack, which utilizes intrinsic data features, comes fairly close on PointNet++, DGCNN, and PointConv, with differences in the range of 2--4\% on the Drop100 attack; it is also much better than the Random drop attack.  
The Proposed approach does not does not make any assumptions about the internal structure of the model. 
Moreover, its performance across different DNN models is fairly consistent; its attack success rates vary by up to 5\% for Drop100 and 13\% for Drop200 across different DNN models. Meanwhile, the Baseline's success rate varies by 20\% for Drop100 across different DNNs, and over 30\% for Drop200. This indicates better generalization of the Proposed method compared to the Baseline approach, even if its success rate is not as high.

\begin{table*}
\centering
\caption{Comparison of the Baseline, Proposed, and Random attacks in terms of attack success rate.}
\label{tab:Attack success rate}      
\tiny
\begin{tabular}{|c|cccccc|}
\hline 
\textbf{Model} &  &\multicolumn{5}{c|}{\textbf{Attacks}} \\ 
\hline
\hline
  &\multicolumn{2}{|c|}{Baseline} & \multicolumn{2}{c|}{Proposed (averaged coefficients)} &\multicolumn{2}{c|}{Random}\\  \cline{2-3} 
\cline{3-5} \cline{5-7}  
  & Drop100~\cite{zheng2019pointcloud} & Drop200~\cite{zheng2019pointcloud} &   Drop100 &  Drop200 & Random Drop100 & Random Drop200 \\

\hline

PointNet\cite{qi2017pointnet}  & \textbf{39.98}  & \textbf{66.97} & 21.80 & 38.13 & 13.82 & 14.10 \\
\hline  

PointNet++\cite{qi2017pointnet1}  & \textbf{19.72}  & \textbf{38.17} & 17.49 & 25.66 & 11.67 & 11.39 \\
\hline  

DGCNN\cite{phan2018dgcnn}  & \textbf{25.46}  & \textbf{43.47} & 22.07 & 35.11 & 10.23 & 11.33 \\
\hline  

PointConv\cite{wu2019pointconv}  & \textbf{22.01}  & \textbf{35.86} & 18.09 & 30.09 & 11.01 & 11.89 \\
\hline  

\end{tabular}
\end{table*}

\subsection{Transferability}
\label{sec:transferability}
When adversarial points are determined for a particular DNN model, it is natural to ask whether the same adversarial points could create a successful attack on another DNN model. The ability to transfer an attack from one DNN model to another 
is an important aspect to consider when evaluating the robustness of an attack strategy. 
In the next experiment, we examine how well the adversarial points for the Drop100 attack identified on one DNN model (``source'') work on another DNN model (``target'').
What this means for the Proposed attack is that the coefficients of the linear model~(\ref{eq:MLR}) are fitted on the source model and then used to predict adversarial drop points on the target model. 

Table~\ref{tab:Transferability} shows the transferability results in terms of the success rate of the Drop100 attack 
on the ModelNet40 test set. 
When the source and target models match, the Baseline approach has a higher success rate than the Proposed method, as expected. However, the success rate of the Baseline approach drops significantly when the target is different from the source. For example, there is more than a 20\% drop in the success rate of the Baseline attack when the source model is PointNet and the target is changed to any of the other three models. 
Meanwhile, the success rate of the Proposed approach also degrades when the source and target models are not the same, but this degradation is much smaller than with the Baseline approach. In fact, the success rate of the Proposed method is higher than the Baseline when the source and target models differ.
Overall, the Proposed approach exhibits higher transferability and a more predictable performance against an unknown target model.

\begin{table*}
\centering
\caption{Transferability of Drop100 attacks in terms of attack success rate.}
\label{tab:Transferability}       
\tiny
\begin{tabular}{|c|c|c|c|c|c|}
\cline{3-6} 
\multicolumn{2}{c|}{}  & \multicolumn{4}{|c|}{\textbf{Target DNN model}}\\
\hline
\textbf{Source DNN Model} & \textbf{Drop100 Attack} & PointNet\cite{qi2017pointnet} & PointNet++\cite{qi2017pointnet1} & DGCNN\cite{phan2018dgcnn} & PointConv\cite{wu2019pointconv} \\
\hline
\hline

\multirow{2}{*}{PointNet\cite{qi2017pointnet}}
& \multirow{1}{*}{Baseline~\cite{zheng2019pointcloud}} &  \textbf{39.98} & 17.91 & 17.00 & 15.61 \\
& \multirow{1}{*}{Proposed (coefficients)} &  25.40 & \textbf{20.12} & \textbf{21.07} & \textbf{20.31} \\
\hline

\multirow{2}{*}{PointNet++\cite{qi2017pointnet1}}
& \multirow{1}{*}{Baseline~\cite{zheng2019pointcloud}} &  14.21 & \textbf{19.72} & 12.27 & 10.36 \\
& \multirow{1}{*}{Proposed (coefficients)} &  \textbf{17.00} & 17.80 & \textbf{16.96} & \textbf{15.45} \\
\hline

\multirow{2}{*}{DGCNN\cite{phan2018dgcnn}}
& \multirow{1}{*}{Baseline~\cite{zheng2019pointcloud}} &  18.47 & 18.97 & \textbf{25.46} & 17.52 \\
& \multirow{1}{*}{Proposed (coefficients)} &  \textbf{20.77} & \textbf{20.85} & 22.07 & \textbf{19.55} \\
\hline

\multirow{2}{*}{PointConv\cite{wu2019pointconv}}
& \multirow{1}{*}{Baseline~\cite{zheng2019pointcloud}} & 11.93 & 12.88 & 15.25 & \textbf{22.01} \\
& \multirow{1}{*}{Proposed (coefficients)} &  \textbf{16.33} & \textbf{16.73} & \textbf{17.66} & 21.61 \\
\hline

\end{tabular}
\end{table*}






\subsection{Performance of various defenses}
Next, we test the performance of several defenses -- SRS~\cite{yang2021adversarial}, SOR~\cite{zhou2019dup}, DUP-Net~\cite{zhou2019dup}, and If-Defense~\cite{wu2020if} -- against the Baseline and Proposed Drop100 attacks. Here, the Proposed attack is based on the averaged significant coefficients from Tables~\ref{tab_pointnet},~\ref{tab_pointnet++}, and~\ref{tab_dgcnn}. The target model is PointConv, which was not used in the computation of coefficients in Tables~\ref{tab_pointnet},~\ref{tab_pointnet++}, and~\ref{tab_dgcnn}. 
Table~\ref{table:defense} shows the classification accuracy of PointConv under different attacks and defenses. 
For reference, the ``No-attack'' column shows the accuracy when there is no attack. 
The lower classification accuracy indicates that the attack is more successful.  
The table shows that defenses are successful to varying degrees, with If-Defense being the most successful in keeping the classification accuracy high.
The classification accuracies under the two attacks are fairly similar, deviating by only 1-4\%. While the Baseline attack performs better under weaker defenses, the Proposed attack proves slightly better under the strongest defenses, specifically DUP-Net and If-Defense.

Overall, the Proposed attack performs comparably to the Baseline attack against four popular defenses, despite relying on the intrinsic characteristics of the point clouds. In contrast, the Baseline attack benefits from having access to internal model information.
This makes it a fairly robust and universal approach for generating adversarial attacks on point cloud analysis models.


  


\begin{table*}
\caption{Performance of various defenses against Drop100 attacks in terms of PointConv classification accuracy; lower classification accuracy indicates a more successful attack.}

\centering
\label{table:defense}
 \scriptsize
\begin{tabular}{|c|c|c|c|c|c|}
  
\cline{3-4} 
\multicolumn{1}{c}{} & \multicolumn{1}{c}{\bf} & \multicolumn{2}{|c|}{\bf Drop100 Attack}\\
\hline
\textbf{Defense} & No-attack &  Baseline & Proposed  \\
\hline
\hline
No-defense & 88.47\% & 77.99\% & 81.91\%\\
\hline
SRS \cite{yang2021adversarial} & 84.98\% & \textbf{65.10\%} & 66.00\% \\
\hline
SOR \cite{zhou2019dup} & 87.26\% & \textbf{69.34\%} & 71.50\% \\
\hline
DUP-Net \cite{zhou2019dup} & 78.69\% & 63.75\% & \textbf{63.53\%}  \\
\hline
If-Defense \cite{wu2020if} & 87.66\% & 84.42\% & \textbf{84.03\%}  \\
\hline

\end{tabular}
\end{table*}

\subsection{Computational cost}

Next, the computational cost of generating the Baseline and Proposed Drop100 attacks is compared. Again, the Proposed attack is based on the averaged significant coefficients from Tables~\ref{tab_pointnet},~\ref{tab_pointnet++}, and~\ref{tab_dgcnn}. The experiment was carried out in a Google Colab environment using a NVIDIA Tesla T4 GPU. 
Table~\ref{table:cost} shows the average run time in seconds per point cloud needed to generate an attack, averaged over the ModelNet40 test set. As seen in the table, the Proposed attack is approximately \textbf{23 times} faster than the Baseline. This is due to its nature, which allows it to generate the attack directly from the point cloud, without having to pass gradients through the target DNN model. 

\begin{table*}
\caption{Average run-time for generating an adversarial point cloud in a Drop100 attack.}

\centering
\label{table:cost}
 \scriptsize

\begin{tabular}{|c|c|}
  
\hline 
\bf Drop100 Attack  & \multicolumn{1}{c|}{\bf Time (s)} \\
\hline
\hline
 Baseline  & $1.41$\\
 
Proposed & \textbf{$0.06$} \\
\hline 

\end{tabular}
\end{table*}

\subsection{Visualizations} 
\begin{figure*}
     \centering
      \includegraphics[width=0.9\textwidth] {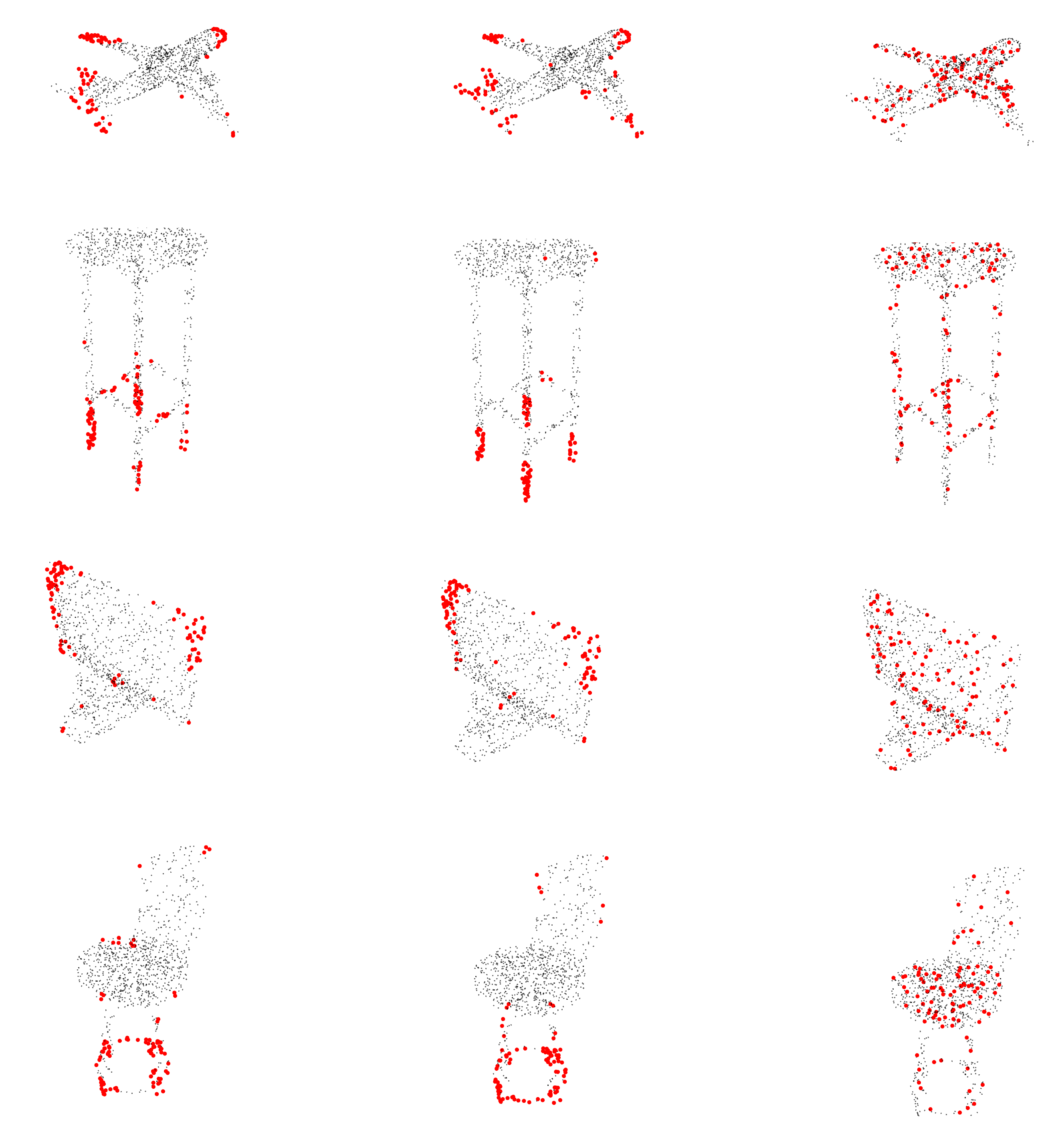}
        \vspace{10pt}
        \caption{Visualisation of adversarial point prediction, where (predicted) adversarial points are shown in red, and the other points in the cloud as gray. (Left) 100 points with the highest true adversarial score computed based on PointNet. (Middle) 100 points with the highest predicted adversarial score computed using the proposed features. (Right) 100 randomly selected points. }
        \label{fig:vis}
\end{figure*}

Finally, several visualizations of true and predicted adversarial drop points are shown in Fig.~\ref{fig:vis}.
The 100 points with the highest adversarial score $z(\p_i)$ from~(\ref{eq:adversarial_score}), computed based on PointNet, are shown in red on the three point clouds on the left, while other points are shown in gray. The middle column shows the 100 points with the highest predicted adversarial score $\mathfrak{z}(\p_i)$ computed from~(\ref{eq:predicted_score}), with the coefficients from Table~\ref{tab_pointnet} for $N=100$.Compared with the red points in the three clouds on the left, there is a very good agreement. For example, true and predicted adversarial points tend to concentrate near the chair legs and edges of the seat (top cloud), tip and sides of the cone (middle cloud), legs and corners (bottom cloud). The last column highlights 100 randomly selected points, which are spread all over the corresponding clouds. Clearly, the predicted adversarial points are much closer to the true adversarial points than a random guess.

\section{Conclusion}
\label{sec:conclude}
In this paper, we explored the key point cloud features that play a critical role in crafting 3D adversarial attacks, shifting the focus from model-specific vulnerabilities to the intrinsic characteristics of point clouds. By defining a set of Fourteen features derived from graph signal processing concepts, features that significantly influence adversarial drop points—points whose removal is likely to alter a model's decision—were identified. Multiple linear regression analysis demonstrated that these features effectively predict adversarial points across several DNN models, including PointNet, PointNet++, DGCNN, and PointConv

The findings highlight that adversarial vulnerability in point clouds stems from their inherent structural properties, rather than merely from flaws in DNNs. Based on these insights, a novel attack method was developed.
While this new attack cannot outperform white- or black-box attacks, as it cannot fine-tune adversarial examples to specific target models, it generalizes better, is more transferable, and is significantly faster than a baseline white-box attack.
One implication of this work is that the geometry of data instances, and not just the architecture of the target model or the data distribution, is important for adversarial attacks, and this insight contribute to the design and development of more effective attacks and defenses in the future.

\section*{Acknowledgement}
\label{sec:acknowledgement}
This work has been supported in part by the Iran National Science Foundation (INSF) and the Natural Sciences and Engineering Research Council (NSERC) of Canada.

\bibliographystyle{elsarticle-num-names} 
\bibliography{main}





\end{document}